\documentclass{article}

\usepackage{arxiv}

\usepackage[utf8]{inputenc} 
\usepackage[T1]{fontenc}    
\usepackage{hyperref}       
\usepackage{url}            
\usepackage{booktabs}       
\usepackage{amsfonts}       
\usepackage{nicefrac}       
\usepackage{microtype}      
\usepackage{lipsum}		
\usepackage{graphicx}
\usepackage{natbib}
\usepackage{doi}
\usepackage{multicol}
\usepackage{multirow} 
\usepackage{caption}
\usepackage{hyperref}
\usepackage{authblk}
\title{InMD-X: Large Language Models for Internal Medicine Doctors}

\author[1]{Hansle Gwon}
\author[1]{Imjin Ahn}
\author[2]{HyoJe Jung}
\author[2]{Byeolhee Kim}
\author[3]{Young-Hak Kim}
\author[4]{Tae Joon Jun}
\affil[1]{INMED DATA, 88, Olympicro 43gil, Songpagu, 05505, Seoul, Republic of Korea}
\affil[2]{Department of Information Medicine, Asan Medical Center, 88, Olympicro 43gil, Songpagu, 05505, Seoul, Republic of Korea}
\affil[3]{Division of Cardiology, Department of Information Medicine, Asan Medical Center, University of Ulsan College of Medicine,  88, Olympicro 43gil, Songpagu, 05505, Seoul, Republic of Korea}
\affil[4]{Big Data Research Center, Asan Institute for Life Sciences, Asan Medical Center, 88, Olympicro 43gil, Songpagu, 05505, Seoul, Republic of Korea}

\hypersetup{
pdftitle={InMD-X: Large Language Models for Internal Medicine Doctors},
pdfsubject={Machine Learning},
pdfkeywords={Large Language Model, medical Language Model, Natural Language Process},
}

\begin{document}
\maketitle

\begin{abstract}
In this paper, we introduce InMD-X, a collection of multiple large language models specifically designed to cater to the unique characteristics and demands of Internal Medicine Doctors (IMD). InMD-X represents a groundbreaking development in natural language processing, offering a suite of language models fine-tuned for various aspects of the internal medicine field. These models encompass a wide range of medical sub-specialties, enabling IMDs to perform more efficient and accurate research, diagnosis, and documentation.
InMD-X's versatility and adaptability make it a valuable tool for improving the healthcare industry, enhancing communication between healthcare professionals, and advancing medical research. Each model within InMD-X is meticulously tailored to address specific challenges faced by IMDs, ensuring the highest level of precision and comprehensiveness in clinical text analysis and decision support.
This paper provides an overview of the design, development, and evaluation of InMD-X, showcasing its potential to revolutionize the way internal medicine practitioners interact with medical data and information. We present results from extensive testing, demonstrating the effectiveness and practical utility of InMD-X in real-world medical scenarios.
\end{abstract}

\section{Introduction}
In recent years, the deployment of large-scale language models has left an indelible mark on the field of healthcare and medicine \cite{thirunavukarasu2023large}. These models have emerged as powerful tools, transforming the way medical professionals analyze and interpret vast amounts of text-based medical data. From assisting in clinical decision-making to automating tedious administrative tasks, the impact of these models on the healthcare landscape has been profound.

The prevalence of large language models in healthcare is undeniable, with numerous models developed and deployed across the medical and healthcare domains \cite{he2023survey}. However, it's worth noting that many of these models have treated healthcare as a monolithic entity, overlooking the intricacies that differentiate medical specialties. A critical limitation of these models has been their inability to discern the unique requirements of various clinical disciplines and the distinct language used in each. This oversight has often resulted in a one-size-fits-all approach, neglecting the specific demands and intricacies of individual medical sub-specialties.

To address this issue, we embarked on a mission to create tailored language models that prioritize the granularity of medical practice. In this paper, we present an innovative approach to fine-tuning large language models, breaking down the field of Internal Medicine into eleven distinct sub-specialties, including general internal medicine. In doing so, we emphasize the significance of differentiating between medical disciplines to provide specialized language models that align with the needs of practicing physicians.

Our endeavor involved meticulous data curation, gathering research papers from top-tier journals in each sub-specialty, and fine-tuning large language models to produce a suite of versatile tools known as InMD-X. This collection of models is poised to revolutionize the way medical professionals interact with and leverage natural language processing capabilities within their specific fields, improving accuracy, efficiency, and relevance across the spectrum of internal medicine (IMD) sub-specialties.
\begin{figure*}[t]
    \centering
    \includegraphics[width=1\columnwidth]{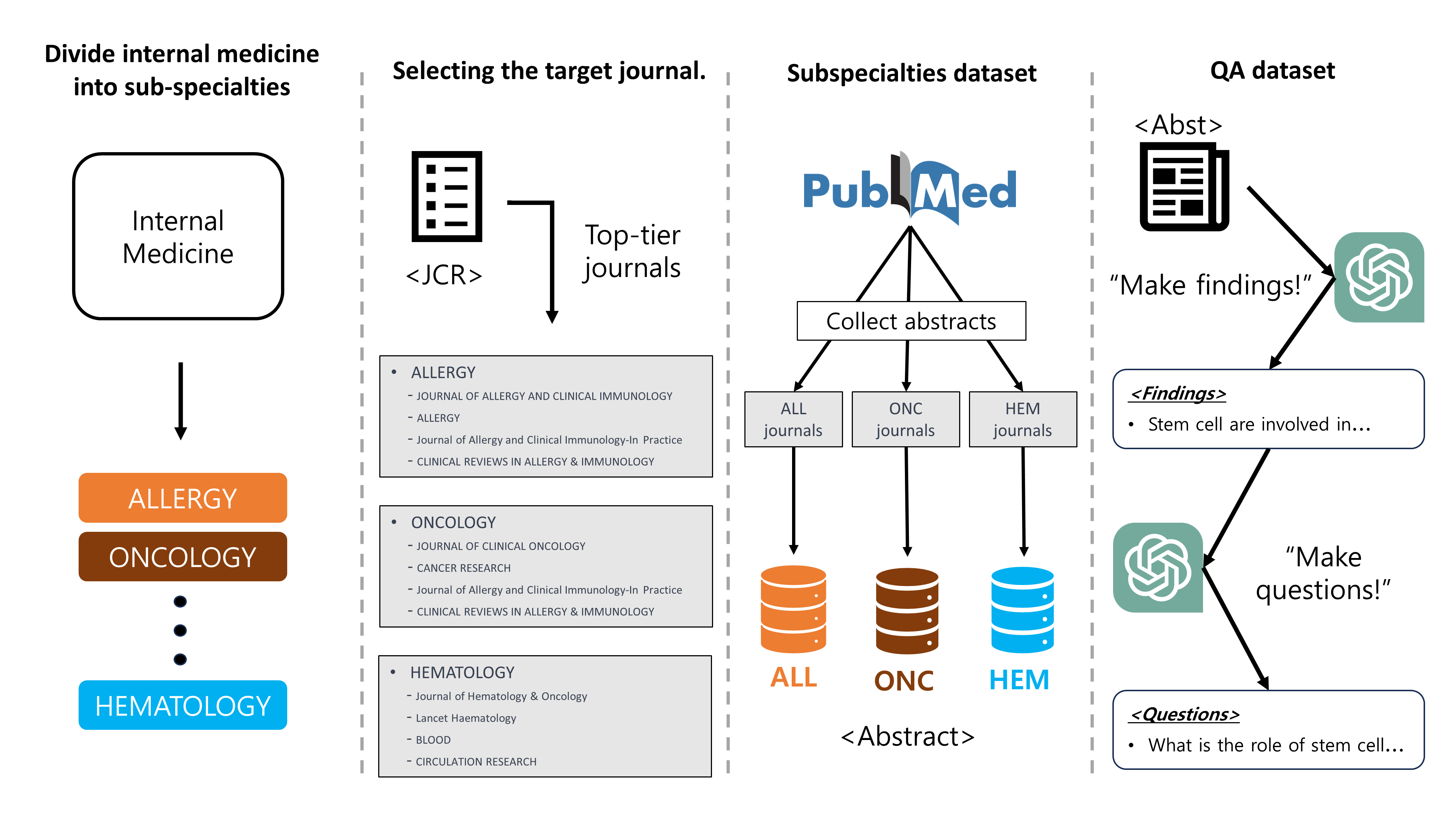}
    \caption{The Dataset Composition Framework of InMD-X.}
    \label{fig:architecture}
\end{figure*}
\section{Related work}

\subsection{Large Language Model}
In recent times, one of the most remarkable successes in the field of AI, particularly within Natural Language Processing, can be attributed to the achievements of Large Language Models (LLMs) such as ChatGPT\cite{achiam2023gpt}. ChatGPT, as an exemplary instance of these LLMs\cite{chowdhery2023palm}\cite{anil2023palm2}, demonstrates exceptional proficiency across a wide spectrum of tasks involving text, ranging from simple question-answering to complex undertakings like generating novels, translation, and even programming tasks. Notably, its potential applicability extends to specialized domains such as healthcare\cite{li2024chatgpt}. The versatility exhibited by ChatGPT distinguishes it from conventional models that primarily focus on a limited set of tasks\cite{vaswani2017attention} \cite{devlin2018bert} \cite{radford2018improving}. The remarkable performance of such LLMs stems from their colossal size and the extensive amount of data used for training. Despite their high performance, these colossal model architectures are not without their limitations, which have drawn the attention of researchers. Firstly, it is worth noting that the companies responsible for developing these LLMs do not disclose their models, rendering them inaccessible to researchers. Even if these models were to be made publicly available, their sheer size poses a significant challenge for ordinary researchers, who often lack the necessary computational resources to effectively work with them, thereby impeding their research efforts.

\subsection{Small Language Model}
The exclusivity and excessively massive model architectures of Large Language Models (LLMs) have restricted accessibility for researchers, prompting the exploration of alternative models that are more accessible. This has led to the emergence of Open Access Small Language Models (SLMs). Llama\cite{touvron2023llama} \cite{touvron2023llama2} stands out as a prominent example of an open-access SLM, demonstrating the ability to achieve performance levels approaching those of LLMs with relatively smaller parameter sizes (ranging from 7 billion to 70 billion). While it should be noted that models like Llama are technically still LLMs, in this research, they are referred to as small language models to emphasize their relative size. The success of such SLMs has spurred research efforts in this direction, leading to the development of various models\cite{chiang2023vicuna} \cite{jiang2023mistral}. However, SLMs are still recognized to have inferior performance compared to LLMs in benchmarks\cite{hendrycks2020measuring} \cite{lin2021truthfulqa} \cite{zellers2019hellaswag} \cite{chollet2019measure}\cite{MMLU} \cite{HellaSwag} \cite{ARC} \cite{TruthfulQA}.

\subsection{Domain specified language model}

Language models are known to scale in their learning capacity and performance capabilities in proportion to the amount of trainable information and the number of parameters. This phenomenon has been a key factor contributing to the significant success of Large Language Models (LLMs) across various domains and remains a major driver for their superior performance compared to Small Language Models (SLMs). One approach to enhancing the performance of SLMs involves the development of domain-specific models, which restrict their learning to a particular domain to achieve high proficiency in that specific field. For instance, Biobert\cite{lee2020biobert} specialized the BERT model for the medical domain and demonstrated impressive performance within the healthcare domain. The success of Biobert has subsequently paved the way for the creation of numerous medical language models, including Meditron\cite{chen2023meditron}, PMC-Llama\cite{wu2023pmc}, ChatDoctor \cite{li2023chatdoctor}, among others. While these research efforts have yielded promising results in the medical domain, they have often treated healthcare as a singular field, overlooking its intricate sub-specialties, each of which demands a high level of expertise.
In this research, our goal is to further refine the medical domain by breaking it down into more granular sub-specialties and constructing models that are specifically tailored to the intricacies of each sub-specialty within the healthcare domain.

\section{Dataset}
In this section, we discuss the methodology employed to construct the dataset utilized for training InMD-X. The dataset construction process involved four distinct phases, and Figure \ref{fig:architecture} illustrates the process of building the InMD-X dataset.

\subsection{Sub-specialties of internal medicine}

While previous research in medical language models has often treated the entire medical field as a single entity, it is crucial to acknowledge that the field of healthcare comprises numerous specialties, each characterized by complex and distinct languages. Even within the domain of internal medicine, there are multiple sub-specialties, each demanding a high level of expertise. In this research, we have redefined internal medicine into 11 distinct sub-specialties. These sub-specialties are as follows: Allergy (ALL), Cardiac Cardiovascular Systems (CAR), Endocrinology Metabolism (MET), Gastroenterology (GAS), Hematology (HEM), Infectious Diseases (INF), Oncology (ONC), Respiratory System (RES), Rheumatology (RHE), Urology Nephrology (URO), and General Internal Medicine (MED).

\subsection{Data Source}

When it comes to training a domain-specific language model, the most crucial aspect is gathering an appropriate domain dataset. We established certain criteria when selecting data sources for dataset construction. (1) There should be a sufficient quantity of data available. (2) The data sources should be open-access databases. (3) Data quality and reliability should be guaranteed. The reasons for setting these conditions are as follows. (1) Training a language model to specialize in a specific domain requires a substantial amount of data.(2) To ensure data accessibility and utilization during both dataset construction and model deployment, open-access databases were deemed necessary. This criterion is relevant not only to minimize costs during dataset construction but also to address licensing issues that may arise when deploying the model after training.(3) In the context of handling medical information, data reliability is one of the most critical requirements. Using inaccurate medical information can hinder appropriate medical practices and research. This concern extends to the training of the language model, as incorrect data can not only degrade model performance but also pose potential risks. 

In accordance with these criteria, we selected Pubmed as the source of our data. Pubmed is an open-access database that provides an extensive collection of research articles in the fields of life sciences, biomedicine, healthcare, and well-being. The data provided by Pubmed undergoes a peer-review process before publication, ensuring high reliability and scholarly value. Additionally, it is authored by medical experts, further enhancing its quality. Pubmed typically provides information such as the title, abstract, journal, and references of research articles. In some cases, it is also linked to Pubmed Central (PMC), allowing access to full-text articles. For our research, we extracted and utilized information from Pubmed, including journal names, titles, abstracts, publication dates, and other relevant data for training purposes.

\subsection{Sub-specialties specified corpus}

Similar to previous research on medical language models\cite{chen2023meditron}\cite{wu2023pmc}, we collected data for training from Pubmed. However, Pubmed does not provide information about specific departments, making it challenging to distinguish and gather data by department. Properly categorizing the collected data has the most significant impact on the model's specialization, and this aspect is considered the most crucial in our research.

We established a two-step process to construct a dataset that clearly distinguishes sub-specialties. In the first step, we selected queries to apply to the Pubmed API. We chose the names of journals that ranked in the top 10\% for each sub-specialty according to the Journal Citation Reports (JCR) 2023\cite{garfield1991journal}. These journal names were used as queries for the Pubmed API.
In the second step, we applied the selected queries to the Pubmed API to extract data. We extracted papers published in the target journals since the year 2010. This method allowed us to collect abstracts from recently published research papers in top-tier journals.

The adoption of this approach is driven by several reasons. Firstly, the Journal Citation Reports (JCR) provides clear departmental classifications. Extracting data based on journals categorized by JCR allows us to collect data accurately for each sub-specialty. For example, research papers published in journals falling under the Cardiac Cardiovascular Systems category in JCR can be considered as a clear corpus related to Cardiac Cardiovascular Systems. Furthermore, since one paper is typically published in only one journal, there is no overlap in data, resulting in clear boundaries between datasets. Secondly, we aimed to use data that medical professionals most frequently read for efficient interaction with physicians. Thus, we extracted journals that belonged to the top 10\% in JCR. Top-tier journals are more likely to feature important research and cater to higher information demand. Thirdly, trends exist in medical research, and over time, conventional concepts and terminology can change. Research papers from too long ago may contain outdated or unnecessary information in contemporary medicine. Therefore, in this research, we selected and used relatively recent research papers published since 2010. The criteria for defining top journals (10\% in this study) or recent publications (since 2010 in this study) can be adjusted according to specific circumstances.

\subsection{QA dataset}

Recent generative language models often employ supervised fine-tuning to align the model with human intent. Supervised fine-tuning involves training the model in a supervised manner to generate desired responses for given inputs. While supervised fine-tuning can yield high performance, it comes with the burden of creating a labeled dataset, where humans specify the desired answers. This task can be particularly challenging in the case of medical models since medical experts are required to create these labels, making it a labor-intensive process. In this research, we propose a method that utilizes the GPT-3.5 API to create a Question \& Answer dataset from the abstracts of medical research papers, without relying on the assistance of medical experts.

To create the Question \& Answer dataset, we initially conducted the process of distilling essential medical knowledge from the extracted abstracts. We refer to the knowledge distilled from the abstracts as "findings". These findings are all single sentences, with typically around 5 findings extracted per abstract. The GPT-3.5 API was utilized for extracting these findings, and the prompt applied during this process was as follows: 

'''
\newline \textit{\textbf{System} : You are a medical AI assistant.}
\newline \textit{\textbf{User} : Read the abstract of the following paper carefully.Identify key findings from medical perspectives step-by-step.}
\newline \textit{Here are requirements:}
\textit{1. Number your findings.}
\newline\textit{2. Start the sentence with a number.}
\newline\textit{3. The finding must not include pronouns.}
\newline\textit{4. Each finding must include at least two medical entities.}
\newline\textit{5. Each finding should be capable of being explained independently, without reference to other findings.}
\newline\textit{6. Refer only to the abstract of the given paper and do not utilize your existing knowledge.}

\textit{\{\underline{\textbf{abstract}}\}}

'''

The findings extracted in this manner sometimes contain information that is not relevant to general medical knowledge. For instance, experiment-specific results and author information that are specific to the given abstract may not be suitable for fine-tuning the model. To address this, we applied a filtering process to the findings. Findings that included keywords such as "study," "paper," "result," "abstract," "author," or "department" were excluded from the training dataset. These filtered findings were then utilized as Answers in the QA dataset.

After completing the extraction of findings, we proceeded to create a dataset of questions derived from these findings. The questions were also constructed as single sentences, and we applied filtering during the question generation process to maintain data quality. We structured the process in this way to focus on findings during the dataset creation. This decision was based on the belief that findings contain practical medical knowledge and would have a more significant impact on model training. Additionally, selecting appropriate questions becomes easier when findings are well-defined. Therefore, we conducted the process in the sequence of finding extraction and filtering followed by the generation of suitable questions.

The cost for finding extraction was approximately 700 \$, and for question generation, it was 427\$, resulting in a total cost of approximately 1127\$ for creating the QA dataset. The question generation process utilized the GPT-3.5 API, and the prompt used was as follows:

'''
\newline\textit{\textbf{System} : You are a medical AI assistant.}
\newline\textit{\textbf{User}:You are a helpful assistant.}
\newline\textit{The following are key findings extracted from the abstract of a paper in PubMed.}
\newline\textit{[Finding]: {\{\underline{\textbf{findings}}}\}.}
\newline\textit{Please write the most appropriate question for the given answer.}
\newline\textit{Here are requirements:}
\newline\textit{1. If the findings are not specific to the biomedical field, respond with 'None'.}
\newline\textit{2. The question must be a single sentence.}

'''

\begin{figure*}[t]
\vskip 0.2in
\begin{center}
\includegraphics[width=1\columnwidth]{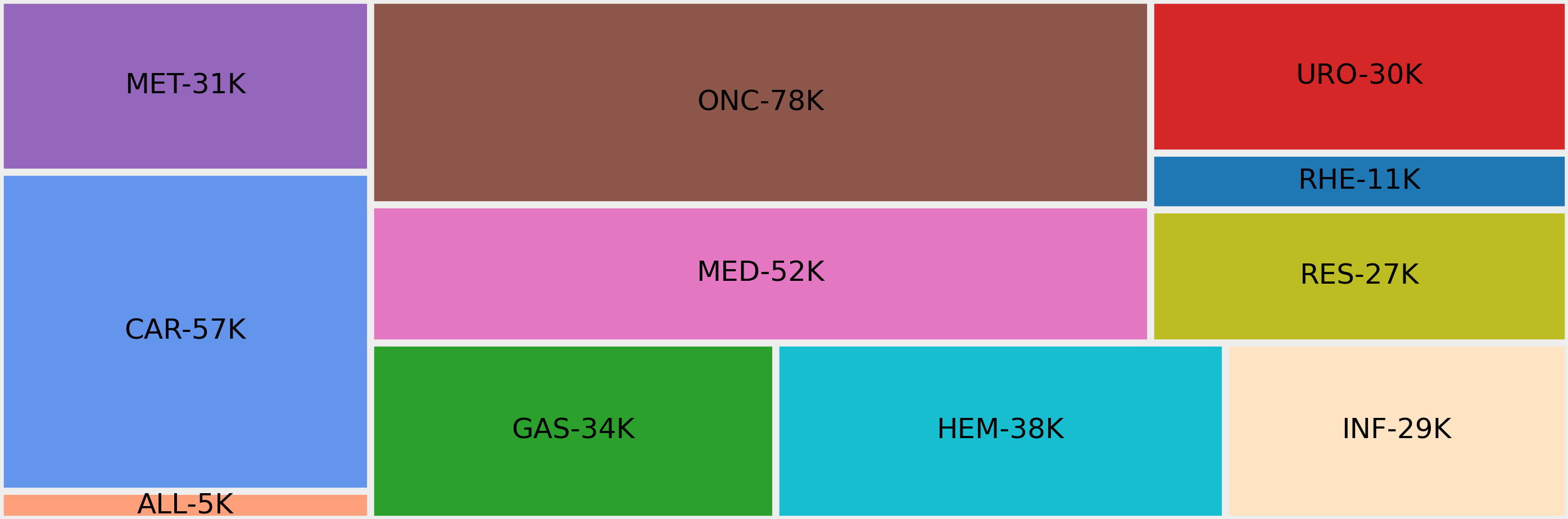}
\caption{The distribution of sub-specialties within internal medicine, based on the number of published papers}
\label{fig:treemap}
\end{center}
\vskip -0.2in
\end{figure*}

\section{Training}

\subsection{Continued pre-training}
The first step in training the InMD-X model involved applying a pre-training process to the pre-trained neural chat-7B language model \cite{neuralchat} to adapt it to the medical domain. In this research, we redefined internal medicine into a total of 11 sub-specialties, and each model was designed to specialize in a single department. Consequently, a total of 11 pre-trained neural chat-7B based models were retrained to suit their respective purposes. The continued pre-training process involved training on the collected research paper abstracts in an auto-regressive manner, without labels. This process enabled the models to gain a high level of expertise in each domain. The continued pre-training lasted for 3 epochs, with a batch size set to 1 and a learning rate of 1e-4. 

\subsection{Supervised fine-tuning}

Following the continued pre-training on the medical dataset, the models underwent supervised fine-tuning. This process involves training the models using a labeled QA dataset to refine their performance in the QA task according to human expectations. The training utilized question-and-answer data extracted from the abstracts. The training was conducted for 1 epoch, with a batch size set to 1 and a learning rate of 1e-4.

\subsection{Parameter efficient fine tuning}

In this research, a total of 11 models were created, one for each sub-specialty within internal medicine. Each of these models had 7 billion parameters, and training all of them simultaneously would be both resource-intensive and inefficient. Therefore, we employed parameter-efficient fine-tuning to optimize the training process. To achieve parameter-efficient fine-tuning, we utilized the Lora framework \cite{hu2021lora}. Lora operates by tuning the model in an indirect manner through separate modules with fewer parameters, rather than training the entire model's weights. Once the training of these modules is completed, their weights are combined with the original model's weights. This integration results in the final weights, which enable the model to better understand the language of the specific domain.

The models trained using parameter-efficient fine-tuning also offer advantages during inference. Our models are highly specialized and, in some cases, it may be necessary to use multiple models simultaneously. The simplest way to use multiple models is to load all of them into memory. However, loading multiple models, each with around 7 billion parameters, can be resource-intensive and sometimes even impractical. An alternative to loading all models simultaneously is to load only the necessary models for a given task and unload them after use. However, repeatedly loading and unloading such large models can consume a significant amount of time, resulting in longer inference times, which is critical in the case of SLMs.

In contrast, models trained using parameter-efficient fine-tuning work by merging Lora weights with the base model. This means that you can load the base model and, depending on the task, load the relevant Lora weights and merge them with the base model to handle multi-model scenarios. Since Lora weights are much smaller in size compared to the model's weights, loading them consumes less time. Furthermore, only one base model needs to be loaded at any given time, minimizing resource usage.

\begin{table*}[t]
\caption{Statistics of the collected dataset. This includes the number of journals used, the total number of papers collected, the average number of papers collected per journal, the number of training tokens, the count of extracted QA pairs, and the number of training steps.}
\label{table:dataset_statics}

\centering
\vskip 0.15in
\begin{center}
\begin{small}
\begin{sc}
\begin{tabular}{|c|c|c|c|c|c|c|c|}
\toprule
      & \textit{\textbf{journals}} & \textit{\textbf{papers}} & \textit{\textbf{papers per journal}} & \textit{\textbf{tokens}} &\textit{\textbf{Q \& A }} &\textit{\textbf{ PT steps }}  & \textit{\textbf{SFT steps }} \\ 
\midrule
\textbf{ALL} & 39           & 5049      & 129.46             & 2.2M   & 19781   & 25K   & 1.1K     \\ 
\textbf{CAR} & 144          & 57810     & 401.45             & 19.5M  & 254924  & 130K  & 15.5K  \\ 
\textbf{MET} & 185          & 31056     & 167.87             & 13.4M  & 146444  & 171K  & 9.0K \\ 
\textbf{GAS} & 140          & 34964     & 249.74             & 13.8M  & 139324  & 73K  & 8.5K\\ 
\textbf{HEM} & 98           & 38816     & 396.08             & 14.5M  & 180980  & 89K  & 11K \\ 
\textbf{INF} & 130          & 29579     & 227.53             & 9.8M   & 125679  & 68K  & 7.5K \\ 
\textbf{MED} & 330          & 52490     & 159.06             & 18.3M  & 214095  & 114K  & 13K \\ 
\textbf{ONC} & 319          & 77977     & 244.44             & 33.0M  & 343642  & 158K  & 21K \\ 
\textbf{RES} & 102          & 27160     & 266.27             & 10.6M  & 113753  & 58K  & 7.0K \\ 
\textbf{RHE} & 56           & 11553     & 206.30             & 4.4M   & 40785   & 25K   & 2.0K\\ 
\textbf{URO} & 126          & 31148     & 247.20             & 11.0M  & 122306  & 68K  & 7.5K \\ 
\textbf{Total}& 1669        & 397602    & 238.22             & 150.6M & 1701713 & 979K & 103.1K   \\ 
\bottomrule
\end{tabular}
\end{sc}
\end{small}
\end{center}
\vskip -0.1in
\end{table*}
\section{Results}

\subsection{Dataset}
In order to construct language models optimized for each of the 11 sub-specialties within internal medicine, we divided internal medicine into 11 departments and constructed datasets accordingly. Following the criteria outlined in Section 2, we selected top-tier journals, and abstracts of research papers published in these journals were collected as training data. Pubmed served as the data source for this endeavor. The total number of selected top 10\% journals amounted to 1669, from which a total of 397,602 papers were extracted. The total number of tokens trained was approximately 150.6 million, and a supervised fine-tuning QA dataset comprising approximately 1.7 million pairs was extracted. Table \ref{table:dataset_statics} provides detailed information regarding the extracted data.

The collected datasets exhibited variations across different medical specialties. Firstly, there were differences in the quantity of journals selected within the top 10\%. The number of journals ranged from a minimum of 39 to a maximum of 330, which indirectly reflects the research activity within each respective field. A higher number of journals may indicate a more active research landscape in that particular specialty. This variability in research activity can result in differences in the volume and diversity of data when constructing datasets from journals.Another notable characteristic that varied among the sub-specialties was the average number of papers published per journal. Some sub-specialties had a concentration of a significant number of papers published in a small number of journals, while others had a diverse range of journals with an appropriate distribution of papers. Assuming that papers within the same journal tend to cover relatively similar content compared to papers from different journals, the number of papers per journal can influence the diversity of the dataset. The distribution of each sub-specialty's representation is depicted in Figure \ref{fig:treemap}.

\subsection{Inference}

\begin{table*}[t]
\caption{Table \ref{table:inference_time} shows the inference time (in seconds) and the number of model/weight loads for 100 rounds of inference. "Full" represents the results when following the conventional approach of loading and unloading the entire model, while "Lora" represents the results when loading only the base model and then loading Lora weights while keeping the base model loaded.}
\label{table:inference_time}
\vskip 0.15in
\begin{center}
\begin{small}
\begin{sc}
\centering
\begin{tabular}{|ll|l|l|l|l|l|}
\hline
\multicolumn{2}{|l|}{}                                                           & \multicolumn{1}{c|}{\textbf{1st}} & \multicolumn{1}{c|}{\textbf{2nd}} & \multicolumn{1}{c|}{\textbf{3rd}} & \multicolumn{1}{c|}{\textbf{4th}} & \multicolumn{1}{c|}{\textbf{5th}} \\ \hline
\multicolumn{1}{|l|}{\multirow{2}{*}{\textbf{Full}}} & \textbf{infer time}       & 630.25                          & 498.25                          & 584.51                          & 547.92                          & 490.98                          \\ \cline{2-7} 
\multicolumn{1}{|l|}{}                               & \textbf{\#\_model\_load}  & 63                              & 47                              & 60                              & 54                              & 47                              \\ \hline
\multicolumn{1}{|l|}{\multirow{2}{*}{\textbf{Lora}}} & \textbf{infer time}       & 312.62                          & 315.46                          & 337.39                          & 314.94                          & 298.78                          \\ \cline{2-7} 
\multicolumn{1}{|l|}{}                               & \textbf{\#\_weight\_load} & 100                             & 100                             & 100                             & 100                             & 100                             \\ \hline

\end{tabular}
\end{sc}
\end{small}
\end{center}
\vskip -0.1in
\end{table*}

In this section, we delve into the experimental results pertaining to inference. First and foremost, we conducted experiments to assess the impact of parameter-efficient learning, one of the primary approaches in this research, on inference time. The experiments were designed to measure inference times under conditions where resources for simultaneously loading all models were limited. For the experiment, we assumed a GPU memory capacity of 40GB and conducted the experiments based on this environment. In this setting, it was possible to load a maximum of 5 models concurrently. To ensure that each sub-specialty model processed only relevant questions, certain conditions were imposed. For instance, questions related to Oncology were processed exclusively by the Oncology model. The experiments were carried out through the following steps:

\textbf{(1)} Load 5 sub-specialty models randomly.

\textbf{(2)} Randomly select one sub-specialty and input relevant questions into the model.

\textbf{(3)} If any of the loaded models from step (1) can handle the question selected in step (2), perform inference immediately.

\textbf{(4)} If none of the loaded models from step (1) can handle the question selected in step (2), randomly unload one of the 5 loaded models and load a new model to process the question, then proceed with inference.

\textbf{(5)} Repeat steps (1) to (4) for a total of 100 iterations and measure the total time taken.

The above process was conducted using two different methods:

- Loading and unloading the entire model (conventional approach).

- Loading the base model once and only loading Lora weights, then combining them with the base model (our approach).

The experimental setup for the experiments was as follows:
\newline
\newline \textbf{GPU} - Nvidia A100(40gb)
\newline \textbf{CPU} - intel i9-14900K
\newline \textbf{python}-3.10.12
\newline \textbf{transformers} - 4.37.0.dev0
\newline \textbf{torch}-2.1.1+cu118
\newline

In the inference process, the "text-generation" pipeline API from the Transformers library was used.

When performing a single inference with the 7B model in the mentioned environment, the time taken was an average of 2.25 seconds, with a minimum of 0.36 seconds and a maximum of 6.40 seconds. Loading the 7B model took approximately 5.0 seconds. The experiments were conducted five times, and the results are presented in Table \ref{table:inference_time}. Using the conventional approach of loading the full model, the total time for 100 inferences ranged from 490.98 seconds to 630.25 seconds. This inference time depends on how many model loads occur. Among the 100 inferences, the scenario with 63 model loads took the longest time, which was 630.25 seconds. On the other hand, using the Lora weight loading approach, the time ranged from 298.78 seconds to 337.39 seconds. The Lora weight loading approach consistently had lower variance because it followed the same procedure for every inference. Based on these results, it can be concluded that the Lora approach is more efficient in terms of total inference time and stability.

Table \ref{table:inference_output} presents the results of inference for two different questions using the neural-chat and InMD-X models. From the results shown in Table \ref{table:inference_output}, it is evident that InMD-X provides more accurate content compared to the baseline model, neural-chat 7B. Additionally, in experiments not shown in Table \ref{table:inference_output}, neural-chat tends to produce more general and longer outputs, whereas InMD-X tends to generate concise and deterministic responses. This aligns with our intention as we trained InMD-X using QA datasets structured as single sentences. While the performance may vary for each question, InMD-X generally appears to provide more effective responses in each sub-specialty.

\begin{table*}[]
\caption{The comparative results between Neural-chat and InMD-X for two specific questions. "Answer" represents the model's response and the closer it is to the labeled answer, the higher the model's performance is considered to be.}
\label{table:inference_output}
\centering
\begin{tabular}{l|ll}
                                                                     & \multicolumn{1}{c|}{{ \textbf{Question 1}}}                                                                                                                                         & \multicolumn{1}{c|}{{ \textbf{Quesiont 2}}}                                                                                                                                            \\ \hline \hline
\textbf{Instruction}                                                 & \multicolumn{2}{c|}{Answer the next question in one sentence.}                                                                                                                                                                                                                                                                                                                  \\ \hline
\textbf{Question}                                                    & \multicolumn{1}{l|}{\textit{\begin{tabular}[c]{@{}l@{}}What were the differences in the \\ incidence of stroke between the \\ on-pump and off-pump groups in the study?\end{tabular}}} & \multicolumn{1}{l|}{\textit{\begin{tabular}[c]{@{}l@{}}What is the potential use of \\ iPSC-derived exosomes in the \\ treatment of Heart Failure?\end{tabular}}}                         \\ \hline
\textbf{\begin{tabular}[c]{@{}l@{}}Answer\\ (findings)\end{tabular}} & \multicolumn{1}{l|}{\begin{tabular}[c]{@{}l@{}}There were no significant differences \\ in the incidence of stroke between \\ the on-pump and off-pump groups.\end{tabular}}           & \multicolumn{1}{l|}{\begin{tabular}[c]{@{}l@{}}iPSC-Derived Exosomes may serve as a \\ novel therapeutic approach for \\ Heart Failure treatment.\end{tabular}}                           \\ \hline
\textbf{neural-chat 7B}                                              & \multicolumn{1}{l|}{\begin{tabular}[c]{@{}l@{}}The incidence of stroke was lower \\ in the off-pump group compared \\ to the on-pump group.\end{tabular}}                              & \multicolumn{1}{l|}{\begin{tabular}[c]{@{}l@{}}They can be used to deliver therapeutic \\ molecules to the heart, promoting cardiac \\ repair and improving heart function.\end{tabular}} \\ \hline
\textbf{InMD-X}                                                        & \multicolumn{1}{l|}{\begin{tabular}[c]{@{}l@{}}The incidence of stroke was similar \\ between the on-pump and off-pump groups.\end{tabular}}                                           & \multicolumn{1}{l|}{\begin{tabular}[c]{@{}l@{}}iPSC-derived exosomes have the potential to be \\ used as a therapeutic approach for Heart Failure.\end{tabular}}                          \\ \hline
\bottomrule
\end{tabular}
\end{table*}

\subsection{Train loss}

\begin{figure*}[t]
\vskip 0.2in
\begin{center}
\includegraphics[width=1\columnwidth]{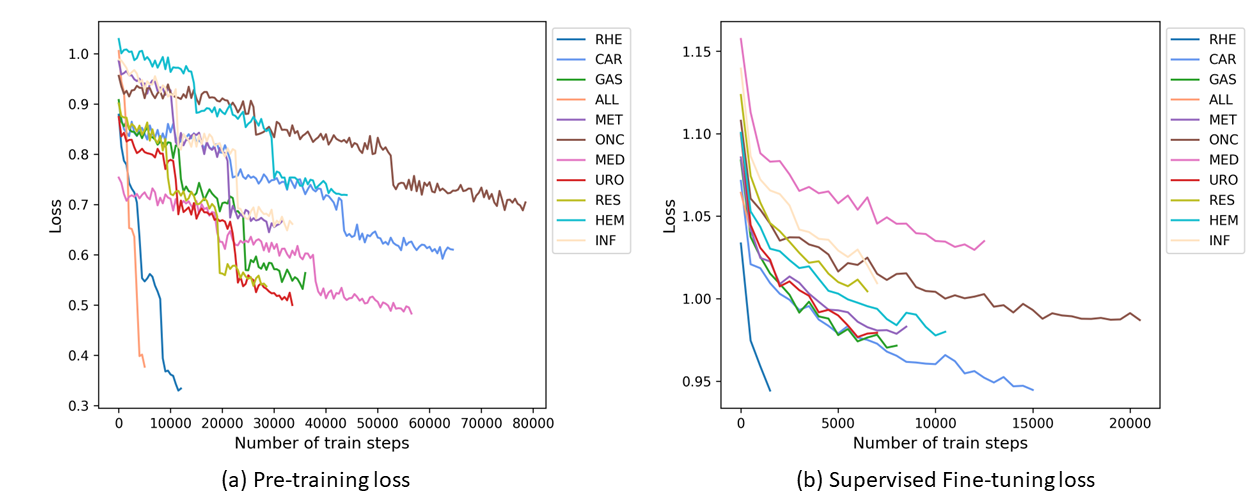}
\caption{The distribution of sub-specialties within internal medicine, based on the number of published papers}
\label{fig:loss}
\end{center}
\vskip -0.2in
\end{figure*}

During the pre-training phase, the training loss was recorded every 500 steps. Training was conducted over a total of 3 epochs. Figure \ref{fig:loss}-(a) displays the training loss graph during the pre-training phase. The training loss exhibited different patterns depending on the type and quantity of the collected data. As expected, the volume of data had the most significant impact on the training patterns. Sub-specialties like Allergy (ALL) and Rheumatology (RHE) had the least amount of training tokens. These two sub-specialties displayed a distinct pattern, rapidly converging compared to other sections. In contrast, Oncology (ONC), which had the most training data, showed a stable and gradual convergence pattern.

In addition to the quantity of data, the diversity of data was found to influence training patterns. Datasets extracted from various journals are assumed to have higher data diversity. Cardiac Cardiovascular Systems (CAR) and General Internal Medicine (MED), while using a similar number of papers, exhibited distinct training patterns. We postulated that this difference stemmed from data diversity. The "papers per journal" metric for Cardiac Cardiovascular Systems (CAR) was 401.45, whereas for General Internal Medicine (MED), it was 159.06. General Internal Medicine (MED) benefited from more diverse data during training, leading to faster convergence during the pre-training phase. A similar pattern was observed in Hematology (HEM), which had the highest "papers per journal" among sub-specialties with approximately 30,000 papers. However, despite having a larger dataset, it exhibited a slower decrease in train loss during pre-training.

During the supervised fine-tuning phase, differences in training patterns based on the data were also observed. Generally, the training patterns showed that more abundant and diverse data led to slower convergence. Firstly, Oncology (ONC), which had the most data used for training, exhibited the second slowest convergence. General Internal Medicine (MED), except for Allergy(ALL), had the most diverse data and showed slow convergence, indicating that a sufficient amount of data contributed to this pattern. This pattern in General Internal Medicine (MED) was compared to Cardiac Cardiovascular Systems (CAR), which trained on a similar amount of data. Cardiac Cardiovascular Systems (CAR) converged quickly despite training on a comparable amount of data, possibly due to a lack of data diversity. Figure \ref{fig:loss}-(b) displays all the learning outcomes of SFT.

\section{Conclusion}

In this research, we proposed InMD-X, a model optimized for each sub-specialty within internal medicine, which is a critical aspect of this study. To achieve this, we redefined internal medicine into a total of 11 sub-categories and collected and processed the dataset in four steps. The data was collected from Pubmed, resulting in a total of 397,602 research papers published in 1,669 journals. The collected data exhibited significant variations among sub-specialties, including differences in data quantity and diversity driven by the number of journals. These differences were reflected in the learning process, where data quantity had the most significant impact, and data diversity also appeared to have some influence.

The training process involved two main steps: pre-training and supervised fine-tuning. Pre-training was performed using an auto-regressive approach on unlabeled data, followed by supervised fine-tuning on labeled QA datasets. Additionally, we applied parameter-efficient fine-tuning with Lora-based techniques, which proved to be suitable for our model, considering the need to train multiple models. 

The primary limitation of our study is the lack of quantifiable evaluation metrics. Since our model further subdivides the medical field, there is no readily available benchmark for its evaluation. We plan to address this limitation in future study, possibly by integrating models using the Mixture of Expert\cite{shazeer2017outrageously} approach and evaluating them on established medical benchmarks like PubMedQA\cite{jin1909dataset}, MedMCQA\cite{pal2022medmcqa}, and MedQA\cite{jin2021disease}. In some cases, creating appropriate benchmarks or considering human evaluations in collaboration with medical experts may also be viable options.

\section{Impact Statements}
This paper presents a method for constructing a more practical medical language model. Our approach, which aligns language models more closely with actual medical systems, offers a new perspective on medical language models. Additionally, the dataset construction method we propose is ethically superior as it preemptively addresses issues of licensing and personal information.


\end{document}